# Demographic Metadata as Construct-Irrelevant Noise in DistilBERT-Based Automated Essay Scoring

Teik Peng Ch'ng
Sunway University
School of Computing and Artificial Intelligence, Faculty of Engineering and Technology, Sunway University, Malaysia
24023731@imail.sunway.edu.my
0009-0000-4511-1116

Hui Na Chua
Sunway University
School of Computing and Artificial Intelligence, Faculty of Engineering and Technology, Sunway University, Malaysia
huinac@sunway.edu.my

## Abstract

Automated Essay Scoring (AES) systems are increasingly used to support teachers in managing grading workloads and to provide a supplementary rater in large-scale assessments. While human grading is frequently influenced by students' demographic characteristics, the efficacy of different strategies for integrating demographic metadata with textual input used to train AES models remains underexplored. This study investigates the impact of a specific multimodal fusion strategy – naïve metadata concatenation - on the predictive accuracy, training convergence, and score parity of a DistilBERT-based AES model. A comparative analysis was conducted using the ASAP 2.0 dataset to evaluate a baseline model against an experimental model trained with input that concatenates tokenised text and demographic metadata using a naive multimodal fusion strategy. Evaluated via 10-fold cross-validation, the findings reveal that the early fusion of demographic metadata and the input significantly degrades the model's overall predictive accuracy. The baseline model achieved a Quadratic Weighted Kappa (QWK) of 0.727, which dropped to 0.656 upon integrating metadata. Furthermore, the experimental model exhibited higher validation loss (1.29) compared to the baseline model (1.25). The experimental model also displayed exacerbated scoring bias, reducing score parity instances from 15 to 12 out of 19 tests.

## 1 Introduction

Automated Essay Scoring (AES) refers to the algorithms or computational models that can predict students' essay scores with a high degree of correlation with human raters' scores (Uto, 2021). AES aims to alleviate teachers' marking burdens in low-stakes classroom assessments rather than fully replace human raters. Additionally, it can act as an additional rater for cross-marking, ensuring the test reliability of large-scale high-stakes assessments (Lim et al., 2021).

However, an investigation into the datasets used for training AES models in prior studies revealed that these datasets primarily consist of numerical essay scores and raw text. This approach assumes that students' writing performance can be assessed based solely on their essays without taking students' demographic backgrounds into account. However, this assumption is challenged by Zanga and De Gioannis (2023), who suggested that teachers' grading is frequently influenced by students' demographic characteristics such as immigration backgrounds, obesity, socio-economic status, ethnicity, gender, special education needs and caste. Therefore, when demographic data is taken into consideration in the process of essay grading, a critical question regarding how AES models should handle these data emerges.

Specifically, it remains unknown whether a naïve multimodal fusion strategy, which directly concatenates demographic metadata with textual input, could improve predictive performance, reduce validation loss and mitigate grading bias. To address this gap, this study aims to evaluate the efficacy and fairness implications of directly concatenating demographic metadata with textual input on the model's predictive performance, training convergence and score parity using the ASAP 2.0 dataset. This matters because AES systems deployed in real educational settings must not inadvertently amplify existing inequities by treating demographic characteristics as scoring signals. To address this, three research questions guide this study:

RQ1: Does naïve concatenation of demographic metadata with textual input improve or degrade the overall predictive accuracy of a DistilBERT-based AES model, as measured by Quadratic Weighted Kappa (QWK) and Linear Weighted Kappa (LWK)?

RQ2: How does naïve metadata concatenation affect training convergence, as measured by validation loss across epochs?

RQ3: Does naïve metadata concatenation improve or worsen score parity across demographic subgroups, as assessed by the Mann-Whitney U test of prediction residuals?.

## 2 Literature Review

Traditional AES models are organised into single-focus and multi-focus architectures. Project Essay Grader (PEG) and Intelligent Essay Assessor (IEA) are examples of the single-focus architecture. PEG, which primarily focuses on syntactic textual features, predicts essay quality by calculating the correlation between the intrinsic essay characteristics and the measurable syntactic features. However, it is susceptible to manipulation as it lacks semantic understanding of an essay (Shermis et. al., 2001). In contrast, the IEA mainly relies on the semantic features of an essay, but fails to assess students' mastery of grammar, which is a core criterion of essay evaluation (Landauer et al., 1998).

The e-rater and the Automated Essay Scoring with Feedback model (AESF) are examples of multi-focus architecture. Both of these models predict an essay's quality using the syntactic and semantic features. However, the performance of the e-rater is restricted by its equal-weight scheme, which assigns equal weights to all microfeatures. Even when alternative models assigning different weights were proposed, predictions showed minimal variation (Chen et al., 2017), suggesting that the underlying limitation lies within the manual features themselves. Similarly, the AESF only relies on a constrained training set of approximately 143 essays and 17 overlapping syntactic features (Ng et al., 2019 ). Therefore, its accuracy is limited by the problems of underfitting and multicollinearity. Additionally, with only 17 syntactic features, the AESF fails to capture the full depth of a student's writing ability. Ultimately, manual feature engineering, which is the core mechanism of single-focus and multi-focus architectures, is inherently bottlenecked by construct under-representation. Hence, the shift to deep learning in the field of AES is necessary.

The limitations of manual feature engineering motivated the adoption of deep learning architectures that learn linguistic patterns directly from textual input. Bidirectional Encoder Representations from Transformers (BERT), which is a pre-trained model that only contains encoder layers, exemplifies this paradigm shift. BERT's architecture yields two advantages. First, its multi-head self-attention layers assign attention weights to every tokenised word, allowing it to resolve the problem of polysemy based on the surrounding contextual information (Vaswani et al., 2017). Second, BERT's feed-forward network, which is present in every encoder layer, allows the model to provide a holistic judgement of an essay. By capturing linear word order in the lower layers, syntactic relationships in the middle layers, and semantic relationships in the final layers, BERT can understand the semantics, syntax and coherence of an essay holistically (Rogers et al., 2020). However, BERT's superior performance comes with computational trade-offs. It contains approximately 110 million parameters, hence requiring substantial GPU memory and training time (Devlin et al., 2019). This encoding model's demanding computational requirements make it highly impractical for deployment at schools because 60% of primary schools, half of lower secondary and a third of upper secondary schools globally have significantly limited ICT allocations (UNESCO, 2025). To address this limitation, Sanh et al. (2020) introduced DistilBERT, a distilled version of BERT that compresses the baseline model by 40% while retaining 97% of its natural language understanding performance and increasing inference speed by 60%. The efficacy of DistilBERT in educational contexts is supported by Megat Norulazmi Megat Mohamed Noor and Mohamad Fazli Mohamad Badauraudine (2025), who evaluated eleven BERT-based models trained on the ASAP dataset for AES tasks using a limited GPU memory size (12GB GPU). They reported that the DistilBERT-based AES model (a QWK of 0.926) outperformed the standard BERT-based model (a QWK of 0.903) under limited GPU memory size because the distilled model could avoid "Out of Memory" errors and learn sufficient patterns from larger batches. This finding shows that under limited GPU memory size, a DistilBERT-based AES model offers an optimal trade-off between predictive accuracy and computational efficiency, making it a highly practical choice for deployment in schools.

Although BERT-based AES models and DistilBERT-based AES models display promising predictive performance, their predictions are susceptible to demographic bias. The majority of existing research on demographic bias in AES falls under the category of post-hoc bias detection, which analyses the score predictions to identify bias after the model is trained. For example, XLNet has been reported to magnify the small grading biases in human scoring and could weakly predict demographic membership based on its hidden states (Kwako & Ormerod, 2024). Moreover, a decline in AES models' predictive accuracy is reported when they are evaluated using essays written by students from different economic statuses (Yang et al., 2024) and cognitive levels (Schaller et al., 2024). While these studies successfully indicated the presence of demographic bias, their diagnostic nature offered no bias mitigation strategies.

Therefore, there is a critical need to explore a priori bias mitigation strategies, which involve manipulating the model configuration to reduce or eliminate bias in score predictions. Based on a review of the AES bias literature, only one study has approximated this approach to date; that study recorded no apparent effect when evaluating AES models by swapping personal names in training essays across different ethnic groups (Muñoz Sánchez et al., 2024). Nonetheless, it is important to note that this study relied on in-text substitution rather than explicitly concatenating demographic metadata with the textual input. The observed lack of bias is likely attributable to this methodological choice as a single textual feature (a name) might provide insufficient demographic context for the AES models to understand the association between linguistic patterns and demographic metadata. Therefore, research investigating the behaviours of transformer-based AES models trained with textual input concatenated with broader demographic metadata through a naïve multimodal fusion strategy is highly warranted.

## 3 Methodology

The pipeline of the methodology is illustrated in Figure 1 below.

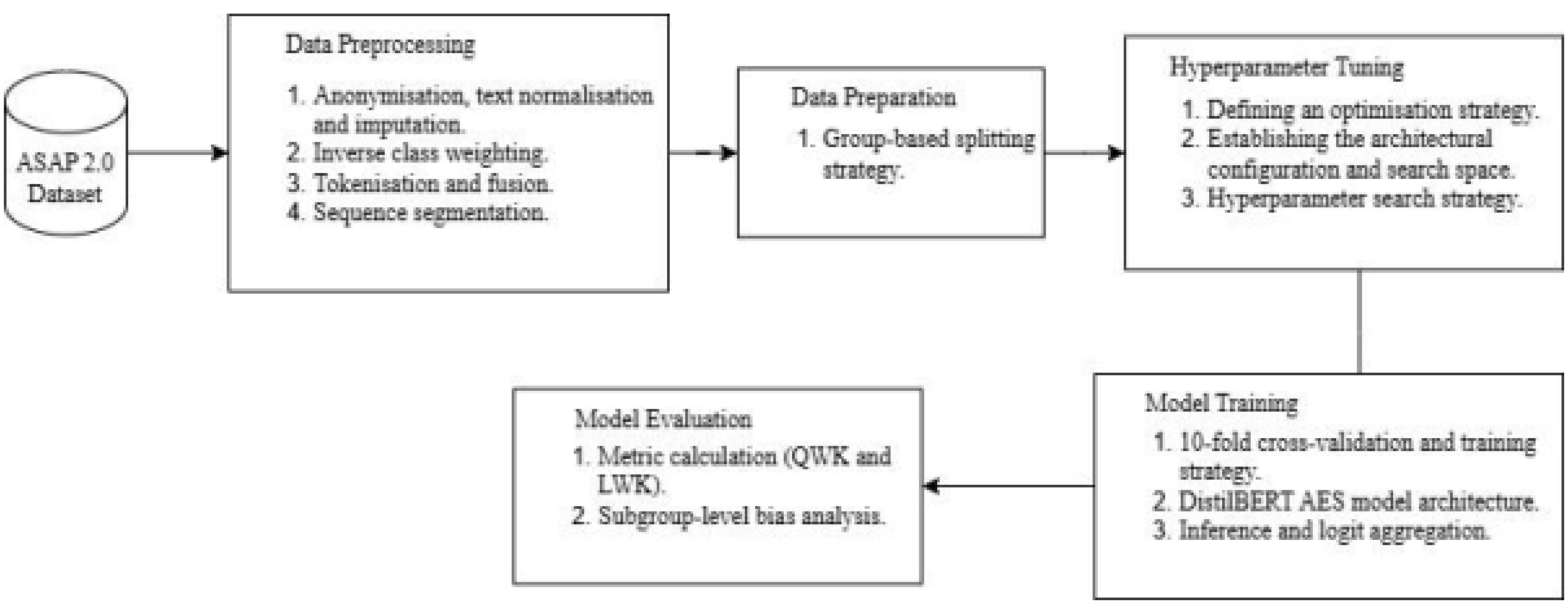

Figure 1: The Pipeline of the Methodology

### 3.1 The ASAP 2.0 Dataset

To empirically evaluate the effects of demographic data on AES performance, this study utilises the Automated Student Assessment Prize 2.0 (ASAP 2.0) dataset. The original ASAP dataset is limited for bias evaluations because it contains exclusively textual input and numerical scores. In contrast, ASAP 2.0 enables bias evaluation because it contains approximately 24,278 essays with comprehensive demographic data, namely economically disadvantaged, students with disabilities, English language learner, gender and race or ethnicity.

The impact of demographic data on the performance of AES models demands empirical investigation because children's language development is closely related to their socioeconomic status (Bahari et al., 2025; Schwab & Lew-Williams, 2016) and cultures (Chunyu, 2025). Additionally, it was reported that teachers' grading is frequently influenced by students' demographic characteristics (Zanga & De Gioannis, 2023). Therefore, the inclusion of demographic metadata in the training of an AES model is insightful as it allows the model to learn and imitate how a human rater grades an essay in different demographic contexts.

## 3.2 Data Preparation

The stage of data splitting includes (i) anonymisation, text normalisation and imputation, (ii) inverse class weighting, (iii) tokenisation and fusion, and (iv) sequence segmentation.

### 3.2.1 Anonymisation, Text Normalisation and Imputation

The personal identifying information (PII) of the texts in ASAP 2.0 corpus was identified and flagged by raters (Crossley et al., 2025). To preserve the context and the syntactic structure of the essays, anonymisation tags were used to replace the de-identified entities.

Additionally, to standardise the data format, text normalisation was performed. For example, the HTML tags were removed, redundant whitespace was collapsed, and curly quotation marks and apostrophes were normalized to their standard ASCII equivalents. To prevent encoding errors caused by multi-byte characters, all raw text data was also decoded using the UTF-8 standard.

Missing data are present in the variables economically_disadvantaged (16.2%), student_disability_status (16.2%), ell_status (1.8%) and race_ethnicity (0.02%). These missing-data rows were not discarded to prevent data loss. Furthermore, common imputation methods - mode, K-Nearest Neighbours (KNN), Random Forest (RF), Sequential Hot-Deck (SHD), and Multiple Imputation by Chained Equations (MICE) - were not performed because their high precision scores do not consistently translate into high predictive accuracy (Memon et al., 2023). NaN was not used to replace missing values because BERT architectures cannot process undefined numerical values. Therefore, missing values were replaced with the descriptive string “Unknown” so that the DistilBERT tokeniser could process missingness as an explicit, recognisable token rather than an undefined value. It is acknowledged that this approach treats missingness as a semantic category, which may introduce its own noise if the model learns spurious associations between the “Unknown” token and score patterns. This is an inherent limitation of token-based representation of missing categorical data and is noted as a direction for future investigation.

### 3.2.2 Inverse Class Weighting

The target variable “score” exhibits class imbalance. For example, Score 3 has 9021 samples while only 200 samples are labelled with Score 6. If the class imbalance is not addressed, the DistilBERT AES model will not have enough samples to learn the linguistic features of Score 6, causing the model to develop a bias towards the majority classes. Therefore, the class weight of each score was calculated so that the model imposed a penalty inversely proportional to class frequency. Hence, errors on the minority classes such as Score 5 and 6 received heavier punishments while errors on the majority classes received lighter punishments. This approach ensured that the model was exposed to a balanced representation of features across the rubric, preventing it from overfitting to the most frequent scores.

**3.2.3 Tokenisation and Fusion**

This study used the pre-trained DistilBERTTokenizer to convert raw strings into integer sequences. Additionally, an early, naive fusion strategy was implemented to keep the sequential input as the only manipulated variable, ensuring a meaningful comparison between the baseline model and the experimental model. This design isolated the root cause, empirically evaluating whether demographic metadata could inherently improve or degrade the QWK of a DistilBERT-based AES model.

In contrast, the intermediate fusion strategy requires architectural modifications such as the insertion of cross-attention layers, gated fusion mechanisms, or dedicated demographic encoders while the late fusion strategy involves processing textual input and demographic metadata through separate models such as a DistilBERT model and a Multi-Layer Perceptron (Willis & Bakos, 2025). These fusion strategies were rejected because the architectural differences between the baseline model and experimental model introduce confounding variables to this study.

**3.2.4 Sequence Segmentation**

Sequence segmentation was performed. The ASAP 2.0 dataset has a mean of 434 WordPiece tokens and the 75th percentile has 532 WordPiece tokens. In other words, 25% of the essays in the dataset are longer than the maximum sequence of 512 tokens. Therefore, a sliding window was created to segment the students' essays into overlapping chunks so that the model can process the entire text without losing context (Abbasi et al., 2025). Specifically, the window size is set to a maximum of 512 tokens with a stride of 128 tokens.

To isolate the independent variable, a global contextual prefix was manipulated across the two models. For the baseline DistilBERT AES model, which was trained only with tokenised textual input, the global contextual prefix was created using only the essay prompts. In contrast, the experimental DistilBERT AES model was trained with tokenised textual input concatenated with demographic metadata. Therefore, its global contextual prefix was created by fusing essay prompts with demographic metadata. The metadata variable was defined in the structure of “key: value” so that the model accurately interpreted and mapped the demographic metadata.

In contrast, for both the models, the complete essay text was divided into overlapping chunks. To comply with DistilBERT’s maximum sequence limit, the length of each chunk was dynamically capped at the length of 512 tokens deducting the length of the global contextual prefix and an additional three tokens. The additional 3 tokens accounted for the specific DistilBERT tokens of [CLS], [SEP], and the final [SEP]. Since this study involved prompt-dependent scoring, the variable window was created by concatenating the global contextual prefix with essay chunks to provide insights into topic relevance and prompt adherence.

After sequence segmentation, some chunks might contain only a few leftover real tokens, followed by hundreds of padding tokens, [PAD]. These padding tokens do not contain semantic meaning or grammatical structures. If they are not removed, the models will extract scoring features from these meaningless tokens, leading to skewed gradients and inaccurate predictions. Therefore, chunks with less than 50 real tokens were filtered out before the stage of data split.

### 3.3 Data Splitting

Because students’ essays were segmented into overlapping chunks using a sliding window, a standard randomised data split could place chunks of the same essay in both the training set and the test set, causing data leakage. Therefore, a group-based splitting strategy was adopted so that essay chunks of the same unique identifiers were treated as indivisible groups and exclusively assigned to either the training set or test set. Data splitting occurred twice in the pipeline. The first data split was performed to divide the dataset into a primary training set and a hold-out test set at a ratio of 80:20 for final model evaluation. The second data split was performed to divide the primary training set into training and validation subsets at a ratio of 75:25 for hyperparameter tuning.

### 3.4 Hyperparameter Tuning

The stage of hyperparameter tuning involved defining an optimisation objective, establishing the architectural configuration and search space as well as developing the hyperparameter search strategy. This tuning stage was performed once on the baseline model to ensure the hyperparameters were optimised for the core task of essay scoring from text, without being influenced by the potentially noisy demographic metadata introduced in the experimental condition. To guarantee a fair comparison, the same set of optimal hyperparameters was utilised to train both the baseline model and experimental model.

### 3.4.1 Defining an Optimisation Objective

Instead of minimising the validation loss, this hyperparameter tuner was configured to optimise the QWK. To maintain computational efficiency, the QWK between the predicted score of each individual chunk and the true score of its parent essay was evaluated at the end of each epoch using the validation subset.

### 3.4.2 Establishing the Architectural Configuration and Search Space

The search space of the hyperparameter tuner was defined as follows:

- The optimal learning rate (lr) was searched within the range of $1e^{-5}$ and $4e^{-5}$ as this range was reported to produce the best performances (Yang et al., 2020).
- The optimal weight decay (wd) was searched within the range of 0.1 and 0.01 to prevent overfitting.
- The optimal dropout rate (dr_rate) was searched within the range of 0.2 and 0.4 with a step size of 0.1.
- With a 12GB GPU, the DistilBERT model could handle up to a batch size of 20 (Megat Mohamed Noor & Mohamed Badauraudine, 2025). This project trained the DistilBERT AES model using NVIDIA L40S, which offers roughly 46GB of VRAM. With more expanded memory capacity, the optimal batch size was searched within the range of 16 and 32.

Additionally, to ensure training stability and mitigate catastrophic forgetting, a learning rate scheduler was adopted to prevent the model from making overly aggressive weight updates early in training when gradient variance was high.

The architecture of the DistilBERT model was constructed using the Keras Functional API. Since foundational English grammar about word order was captured in the lower layers of the DistilBERT model (Rogers et al., 2020), the first two layers were frozen so that the original weights of the pretrained model were not overwritten by weight updates derived from the relatively small ASAP 2.0 dataset. The remaining layers were left trainable so that the model learned the deeper, task-specific features such as contextual, semantic and syntactic information from the dataset for accurate grading.

Following the final transformer block, the output token embeddings were aggregated into a single vector via a global average pooling operation. This architectural choice was adopted to align with the 6-point holistic rubric of the ASAP 2.0 dataset (Crossley et al., 2025). By averaging the contextualised token embeddings, the model captured the overall, aggregate syntactical and semantic quality of the essays. Subsequently, the averaged vector was passed through an optimised dropout layer and finally a fully connected layer consisting of 5 units.

AdamW was adopted as the optimisation function because it shows better generalisation through decoupling the optimal weight decay from the adaptive gradient updates (Loshchilov & Hutter, 2019). Given the ordinal nature of the essay scores, this model was trained using ordinal cross entropy as the loss function because it treats the AES task as discrete classification, imposing proportionately heavier penalties on larger prediction errors while smaller prediction errors receive a lighter penalty (Beckham & Pal, 2017).

#### 3.4.3 Hyperparameter Search Strategy

To identify the optimal hyperparameters using limited computational resources, the Successive Halving Algorithm (SHA) was adopted. During the tuning process, this strategy systematically evaluated randomly sampled configurations, terminating underperforming models early while allocating additional training epochs to promising candidates. This elimination process was iterated across a predefined number of brackets to isolate the single best-performing hyperparameter configuration.

### 3.5 Model Training

This section details the complete model training pipeline, encompassing the 10-fold cross-validation and training strategy, the DistilBERT-AES architecture, and the final inference and logit aggregation procedures.

#### 3.5.1 10-Fold Cross-Validation and Training Strategy

The 10-fold cross-validation strategy was adopted. To prevent data leakage, stratified group-based partitioning was utilised to exclusively assign essay chunks sharing the same unique identifiers to either the training fold or the validation fold. Furthermore, the baseline model and the experimental model were trained using the same set of optimal hyperparameters for fair comparison: a learning rate of $1.328 \times 10^{-5}$, a batch size of 16, a weight decay of 0.01, and a dropout rate of 0.3.

During the training process, tokenised textual input was concatenated with demographic metadata so that this model learned and extracted reliable patterns between essay quality and students' demographics. Given the adoption of the 10-fold cross-validation approach, this training process yielded 10 DistilBERT-AES models, each evaluated using a different validation fold. In subsequent sections, these 10 models are referred to as fold models.

### 3.5.2 DistilBERT AES Model Architecture

The architecture of this AES model in the stage of model training remained consistent with the model in the stage of hyperparameter tuning. During the training loop, a learning rate scheduler was introduced to ensure training stability and mitigate catastrophic forgetting. An early stopping mechanism was also configured to monitor the validation loss at the end of each epoch, triggering early stopping to prevent overfitting and saving the optimal model weights for the subsequent inference stage. Following the forward pass, the aggregated vector was passed through the optimal dropout layer and a fully connected layer consisting of 5 units to yield a chunk-level prediction of 5 threshold logits.

### 3.5.3 Inference and Logit Aggregation

During the inference stage, all the chunk-level predicted logits of each unique essay were concatenated and averaged across the chunk dimension, yielding an array of five raw logits representing the overall quality of the complete essay. Subsequently, a sigmoid activation function was applied to these averaged raw logits, converting them into five threshold probabilities. These threshold probabilities were summed, and the total probability was shifted by a constant of 1 to align with the base index of the grading scale. The resulting value was then rounded to the nearest integer and strictly bounded within the 1-to-6 ordinal scoring rubric to derive the final discrete prediction. Finally, the metrics of Quadratic Weighted Kappa (QWK) and Linear Weighted Kappa (LWK) were calculated by comparing the final discrete predictions against the ground-truth human scores.

## 3.6 Model Evaluation

To evaluate the final generalisation capability of the architecture, an ensemble approach was employed on the unseen test set. The ten DistilBERT fold models generated during the cross-validation phase each produced chunk-level raw logits. To maximize model stability and predictive accuracy, these outputs were ensembled by calculating the mean of the raw logits across all ten fold models. These ensembled chunk-level logits were subsequently aggregated and converted into final discrete essay scores following the identical inference procedure detailed in the Inference and Logit Aggregation section. Finally, the ultimate performance of the ensemble model on the test set was quantified by calculating the Quadratic Weighted Kappa (QWK) and Linear Weighted Kappa (LWK) against the ground-truth human scores.

# 4. Results and Discussions

## 4.1 Impacts of Demographic Metadata on the DistilBERT AES Model

This study constructed two DistilBERT AES Models: (a) The baseline model (trained with only tokenised textual input), and (b) the experimental model (trained with input concatenating tokenised textual input with demographic metadata).

While Muñoz Sánchez et al. (2024) reported that input concatenated with names from different ethnic groups had no significant effect on the results of the BERT-based AES model, the results of this study indicated that concatenating demographic metadata with tokenised textual input degraded the QWK and LWK scores of the DistilBERT AES model. The baseline model achieved a QWK of 0.727. After the inclusion of demographic metadata, its QWK score dropped by 0.071 to 0.656. To contextualise this drop, the QWK of human inter-rater agreement lies within the range of 0.61 and 0.85 (Shermis, 2014). It suggests that the baseline model demonstrated a better alignment with human-level grading reliability than the experimental model. The decrease of predictive accuracy is also observed in the metric of LWK. After the integration of demographic metadata, the baseline model's LWK deteriorated by 0.083, from 0.564 to 0.481.

The decrease of predictive performance suggests that demographic metadata are construct-irrelevant features, which lack significant correlation with the essay scores assigned by human raters. Neural networks are pattern-matching algorithms (Mah & Chakravarthy, 1992). When the DistilBERT AES model received the input concatenated with demographic metadata, it might try to establish spurious mathematical correlations between demographic features and human scores, leading to overfitting.

Additionally, the inclusion of demographic metadata increased the number of features, leading to a sparse feature space. The high dimensionality rendered nearest neighbour search (NNS) through distance measurement meaningless. Consequently, the model failed to identify reliable patterns from the limited data, leading to poor generalisation (Peng et al., 2025).

### 4.2 Shannon Entropy Analysis

The negative impacts of demographic metadata in this study could also be attributed to modality mismatch. DistilBERT is a transformer-based masked language model pre-trained on textual data. The input used to train the experimental model was a combination of categorical demographic metadata and sequential textual data, creating a distributional mismatch between structured and unstructured modalities that may disrupt the scoring rules in the classification head (Billa, 2026). To validate the presence of modality mismatch, Shannon entropy was computed over the discrete probability distributions generated by the classification head. The experimental model produced a higher mean entropy (1.5041) than the baseline (1.4781), and this difference was statistically significant ($p < 0.001$). While the absolute difference is small, the consistent directional pattern across all test folds supports the interpretation that concatenated categorical metadata increased predictive uncertainty in the classification layer. This finding should be interpreted as corroborating evidence for modality mismatch rather than a standalone measure of degradation.

### 4.3 Impacts of Demographic Metadata on the DistilBERT AES Model's Training Convergence

Both models achieved their optimal convergence at epoch 2. The baseline model reached a minimum validation loss of around 1.25 while the experimental model plateaued at a higher minimum of 1.29. Subsequent training epochs resulted in an increase in validation loss for both models, leading to overfitting.

The higher validation loss of the experimental model is consistent with its degraded predictive accuracy reported in the previous section. The inclusion of demographic metadata exposed the model to construct-irrelevant noise, which interfered with the extraction of reliable linguistic patterns. The additional features also increased dimensionality, producing data sparsity that further hindered generalisation. Third, the modality mismatch at the classification head disrupted the model's scoring rules, compounding the increase in validation loss alongside the drops in QWK and LWK.

### 4.4 Impacts of Demographic Metadata on the Score Parity across Subgroups

For the purpose of this analysis, parity was defined as the absence of a statistically significant difference in prediction residuals between demographic subgroups, as determined by a Mann-Whitney U test at the $\alpha = 0.05$ significance level. The parity results of both the baseline model and the experimental model are summarised in Table 1 below.

Close inspection of Table 1 revealed that the baseline model reported bias in two features: (a) student disability status, and (b) race ethnicity, which involves the catch-all class of “two or more races/other”. By comparison, the incorporation of demographic metadata into the input compromised the experimental model’s performance, leading to bias reported in all demographic features, as shown in Table 2.

Table 1: The Baseline Model’s Parity Test Results

| Demographic Feature | Subgroup A | Subgroup B | Baseline Model | U-Statistic | p-value |
|---|---|---|---|---|---|
| Socioeconomic Status | Economically disadvantaged | Not economically disadvantaged | Parity | 2029153 | 0.955029 |
| Disability | Not identified as having disability | Identified as having disability | Biased | 1006573.5 | 0.001423 |
| ELL Status | No | Yes | Parity | 1377131 | 0.314958 |
| Gender | Female | Male | Parity | 2983548 | 0.103936 |
| Race/ Ethnicity | Hispanic/Latino | Two or more races/Other | Biased | 224310 | 0.016699 |
| | White | Two or more races/Other | Biased | 317994 | 0.013606 |
| | Black/ African American | Two or more races/Other | Biased | 146686.5 | 0.006225 |

Table 2: The Experimental Model's Parity Test Results

| Demographic Feature | Subgroup A | Subgroup B | Experimental Model | U-Statistic | p-value |
|---|---|---|---|---|---|
| Socioeconomic Status | Economically disadvantaged | Not economically disadvantaged | Biased | 2117269 | 0.010831 |
| Disability | Not identified as having disability | Identified as having disability | Biased | 963605 | 0.000001 |
| ELL Status | No | Yes | Biased | 1333130 | 0.015407 |
| Gender | Female | Male | Biased | 2956641 | 0.028939 |
| Race/ Ethnicity | Hispanic/Latino | Two or more races/Other | Biased | 223211.5 | 0.029164 |
| | White | Two or more races/Other | Biased | 314270 | 0.041467 |
| | Black/ African American | Two or more races/Other | Biased | 146592.5 | 0.007959 |

The bias indicated in the baseline model's predictions of the student disability status was likely attributable to class imbalance. This dataset contains 17746 sample essays from students without disability and 2977 sample essays from students with disability. Due to the underrepresentation, this model failed to learn sufficient meaningful linguistic patterns of the minority group, resulting in poor generalisation to the unseen test data.

Additionally, the divergent outcomes within the race/ethnicity feature are noteworthy. The bias reported in the baseline model's predictions for the catch-all class "two or more races/other" is likely caused by internal class heterogeneity. This catch-all class aggregates individuals who identified as cross-ethnic groups (e.g. Asian and White, or Black and Hispanic) or unlisted ethnicities during the demographic survey. Therefore, the sample essays of this class exhibit a highly diverse mixture of linguistic patterns. Without a coherent representation of linguistic patterns, the model failed to extract reliable linguistic patterns that could generalise to the unseen test data.

However, once demographic metadata were concatenated with textual input, demographic features (socioeconomic status, ELL status and gender) that achieved parity in the baseline model yielded biased results. The systemic bias observed in the experimental model demonstrated that the metadata introduced structural noise that incentivised the hidden layers to weigh demographic classifications over actual linguistic competence.

### 4.5 Error Analysis

An error analysis contrasting the predictions of the baseline and experimental models indicated that the largest errors made by both models were two rubric units away from the accurate predictions (true scores). This limited error range demonstrated the effectiveness of the ordinal cross-entropy loss function. However, the distribution of these errors across demographic subgroups was obscured.

To further investigate how these errors affect specific demographic subgroups, a subgroup-level mean prediction shift analysis was conducted. The analysis compared the mean human scores against the mean predictions of both models across the demographic subgroups. The results were displayed in Table 3 below.

Table 3: The Mean Prediction Shifts across Demographic Subgroups

| Subgroup | Mean Human Score | Mean Baseline Prediction | Mean Experimental Prediction | Human – Baseline Shift | Baseline – Experimental Shift |
|---|---|---|---|---|---|
| Identified as having disability | 2.36 | 2.56 | 2.76 | +0.20 | +0.20 |
| Not identified as having disability | 3.03 | 3.13 | 3.26 | +0.10 | +0.13 |
| Female | 3.06 | 3.16 | 3.29 | +0.10 | +0.13 |
| Male | 2.89 | 3.00 | 3.14 | +0.11 | +0.14 |
| Not ELL | 3.03 | 3.14 | 3.27 | +0.11 | +0.13 |
| ELL | 2.51 | 2.65 | 2.82 | +0.14 | +0.17 |
| Not economically disadvantaged | 3.17 | 3.28 | 3.39 | +0.11 | +0.11 |
| Economically disadvantaged | 2.77 | 2.89 | 3.06 | +0.12 | +0.17 |
| American Indian/Alaskan Native | 3.19 | 3.25 | 3.38 | +0.06 | +0.13 |
| Asian/Pacific Islander | 3.32 | 3.39 | 3.52 | +0.07 | +0.13 |
| Black/African American | 2.73 | 2.87 | 3.02 | +0.14 | +0.15 |
| White | 3.07 | 3.20 | 3.32 | +0.13 | +0.12 |
| Hispanic/ Latino | 2.85 | 2.98 | 3.12 | +0.13 | +0.14 |
| Two or more races/Other | 3.09 | 3.11 | 3.26 | +0.02 | +0.15 |

The results revealed that the baseline model exhibited an upward shift when compared to true human scores. This baseline inflation was likely attributable to regression to the mean, where the model artificially pulled the predictions of lower-scoring minority groups closer to the global average due to class imbalance.

Additionally, this baseline drift was not exclusively confined to marginalised subgroups, because a mix of both marginalised and non-marginalised subgroups exhibited the highest shifts. For example, White students, the traditionally non-marginalised group, are among the subgroups with the highest shifts, together with other marginalised groups such as Hispanic/Latino students and economically disadvantaged students. In contrast, American Indian/ Alaskan Native students, who were historically marginalised, received only a shift of +0.06. In short, the baseline drift did not exhibit a systemic bias.

In contrast, the experimental model demonstrated a clear trend of systemic bias, as historically marginalised groups—namely students with disabilities, economically disadvantaged students, English language learners, minority groups (Two or more races/Other), Black/African American students, and Hispanic/Latino students were among the subgroups with the most pronounced score inflation. This finding suggested that, despite identical architectural configurations, the naïve concatenation of demographic metadata introduced a systematic upward bias to the internal self-attention layers of the experimental model, causing it to disproportionately favour these specific demographic subgroups.

To illustrate the pattern of systemic bias identified above, a single essay (essay id: AAAOPP13416000002731) accompanied by the 'economically disadvantaged' metadata token is examined as a qualitative example. The prompt is Seagoing Program Experience, and the true score is 2. The baseline model correctly assigned a score of 2, consistent with the essay's characteristics as a beginning writer: simple vocabulary ("good experience," "different places"), repetitive sentence structures ("It's like nothing else" appearing twice in close proximity), and common grammatical inconsistencies ("some people don't like do get out much"). The experimental model assigned a score of 3. This single example is offered as an illustration of the upward inflation pattern identified in Table 3, not as evidence in itself. The systematic nature of this inflation is demonstrated by the subgroup-level mean shift analysis across all demographic groups, not by individual cases.

## 5. Conclusion and Future Work

This study reveals that incorporating demographic metadata significantly degrades the model's overall predictive accuracy. The baseline model achieved a Quadratic Weighted Kappa (QWK) of 0.727 and a Linear Weighted Kappa (LWK) of 0.564. Upon concatenating the textual input with demographic metadata through a naïve fusion strategy, the QWK and LWK of the experimental model decreased to 0.656 and 0.481 respectively. Furthermore, the inclusion of demographic metadata also deteriorated the training convergence of the experimental model while exacerbating scoring bias. These combined results indicate that, for a DistilBERT AES model that adopts a naïve fusion strategy, within this specific hyperparameter configuration, demographic metadata acts as construct-irrelevant noise that introduces a sparse feature space and modality mismatch, ultimately disrupting the AES model's ability to extract and generalise reliable linguistic patterns.

This research makes two contributions to the field of educational NLP. First, it provides the first empirical evaluation of naïve demographic metadata concatenation in a DistilBERT-based AES model trained on the ASAP 2.0 dataset, demonstrating that this strategy degrades both predictive accuracy and fairness simultaneously. Second, and more critically for educational equity, the findings show that naïve metadata concatenation does not merely fail to help — it actively introduces systematic upward score inflation disproportionately affecting historically marginalised groups including students with disabilities, English language learners, and economically disadvantaged students. This is a meaningful warning for practitioners and policymakers considering demographic-aware AES deployment: without architecturally appropriate fusion strategies, incorporating demographic data risks amplifying existing educational inequities rather than reducing them. These findings do not preclude the potential utility of demographic metadata in more sophisticated AES architectures.

While this study provides a foundational understanding of naïve early fusion in AES, several directions for future work are identified. First, despite the application of inverse class weighting, class imbalance continued to affect subgroups such as disability status. Future studies should explore targeted data augmentation or advanced resampling techniques to ensure sufficient representation of minority groups in training. Second, the subgroup-level mean shift analysis (Table 3) identified disability status (+0.20), ELL status (+0.17), and economic disadvantage (+0.17) as the subgroups with the most pronounced score inflation in the experimental model. Future work should conduct feature-level ablation experiments, removing one demographic variable at a time, to identify which specific metadata features drive the observed bias and inform targeted mitigation. Third, to address the modality mismatch documented in this study, future research should evaluate advanced fusion architectures including intermediate fusion with cross-attention layers, gated fusion mechanisms, and adapter modules that inject demographic signals without disrupting the transformer's linguistic processing pipeline (Willis & Bakos, 2025). The immediate next empirical step is a direct comparison of these three fusion strategies on the same ASAP 2.0 dataset and hyperparameter configuration used here, to establish whether any approach can recover the predictive and fairness performance lost by naïve concatenation.

## Statements and Declarations

The authors declare no potential conflicts of interest with respect to the research, authorship, and/or publication of this article, and receive no financial support for its research or publication. The dataset analysed during the current study, "ASAP 2.0: Automated Student Assessment Prize," is publicly available on Kaggle and can be accessed at the following repository: https://www.kaggle.com/datasets/lburleigh/asap-2-0. Furthermore, this research did not involve human or animal subjects; therefore, formal ethical approval was not required.

Generative AI (Gemini) was used as a supportive tool in the preparation of this manuscript. Gemini was employed to assist in the construction, optimization, and debugging of the Python code used for implementing the DistilBERT-based Automated Essay Scoring model. Additionally, Gemini was utilised for fact-checking and verifying technical definitions throughout this manuscript to ensure accuracy. To maintain the academic integrity and authorship of this research, all suggestions generated by Gemini were manually verified for correctness and integrated with original scholarly literature.